\documentclass[10pt]{article}

\usepackage[preprint]{tmlr}
\usepackage{url,hyperref,microtype,subcaption}
\usepackage{amsthm,amsmath,amssymb}
\usepackage{algorithm}

\usepackage{algorithmic}
\usepackage{tikz}
\usepackage{graphicx}
\usetikzlibrary{decorations.text, arrows.meta, positioning}
\usetikzlibrary{decorations.pathreplacing,calc,shapes,positioning}

\theoremstyle{definition}
\newtheorem{example}{Example}
\newtheorem{definition}{Definition}

\usetikzlibrary{shapes,shadows,arrows,positioning,angles,quotes}
\tikzset{My Arrow Style/.style={single arrow, fill=black!15, anchor=base, align=center,text width=2.3cm}}
\tikzstyle{arrow} = [thick,->,>=stealth]
\usetikzlibrary{calc,trees,positioning,arrows,fit,shapes,calc}

\tikzstyle{startstop} = [rectangle, rounded corners, minimum width=1.5cm, minimum height=0.5cm,text centered, draw=black, fill=red!30]
\tikzstyle{io} = [trapezium, trapezium left angle=70, trapezium right angle=110, minimum width=0.5cm, minimum height=0.5cm, text centered, draw=black, fill=blue!30]

\tikzstyle{process} = [rectangle, minimum width=3cm, minimum height=0.5cm, text centered, draw=black, fill=orange!30]
\tikzstyle{decision} = [diamond, minimum width=0.5cm, minimum height=0.1cm, text centered, draw=black, fill=green!30]
\tikzstyle{process2} = [rectangle, minimum width=1cm, minimum height=0.5cm, text centered, draw=black, fill=orange!30]
\tikzstyle{arrow} = [thick,->,>=stealth]

\title{Gimitest: A Comprehensive Tool for Testing Reinforcement Learning Policies}

\author{\name Dennis Gross \email dennis@artigo.ai \\
        \name Quentin Mazouni \email quentin@simula.no \\
        \name Helge Spieker \email helge@simula.no \\
        \name Arnaud Gotlieb \email arnaud@simula.no \\
        \addr Simula Research Laboratory, Oslo, Norway}

\begin{document}

\maketitle

\begin{abstract}
Reinforcement learning (RL) policies can be unsafe and vulnerable to attacks.
Ensuring their reliability is often a pain point as existing automated testing methods target only selected environments, testing scenarios, and RL algorithms. 
To address this, we propose a comprehensive framework for testing single- and multi-agent RL policies under varying conditions.
Our implementation of this framework, \emph{Gimitest}, is an open-source tool that supports various gym frameworks and allows for modifications of their integrated components.
This article describes the framework and details \emph{Gimitest's} functionality and architecture.
It showcases its effectiveness in testing multiple RL policies in environments such as the official Farama Gymnasium and~PettingZoo.
\end{abstract}

\section{Introduction}
\emph{Reinforcement Learning (RL)} has improved various industries~\citep{DBLP:journals/cor/CuiY24,DBLP:journals/eaai/TianYJ24,DBLP:journals/isci/ZhouWZCJ23}, enabling the creation of agents that can outperform humans in sequential decision-making tasks~\citep{mnih2015human}.

\begin{figure}[th]
\centering

    \begin{tikzpicture}[]
     {};
    \node (agent1) [process] {RL Agent $\pi(o)$};
    \node (env) [process, below of=agent1,yshift=0.3cm,xshift=4cm] {Environment};
    
    \draw [arrow] (agent1) -| node[anchor=west] {$a$} (env);
    \draw [arrow] (env) -| node[anchor=east] {$o = \mathbb{O}(s),rew$} (agent1);
    \end{tikzpicture}

\caption{Single-agent RL system, where an agent receives an observation $o$ and a reward $rew$ from the environment after an action $a$, and chooses the next action.}
\label{fig:rl}
\end{figure}
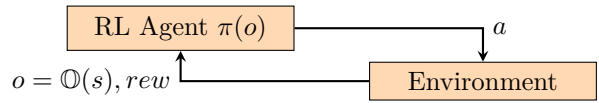

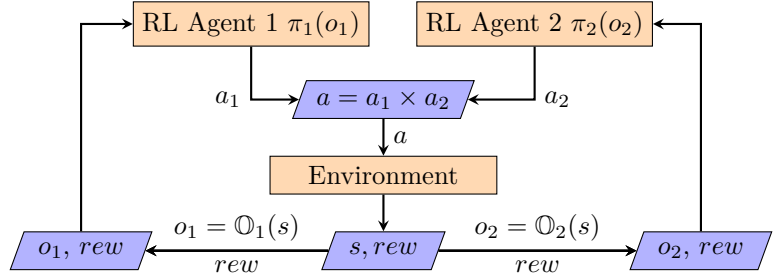
\begin{figure}[th]
\centering

 \begin{tikzpicture}[]
    \node (agent1) [process] {RL Agent 1 $\pi_1(o_1)$};
    \node (agent2) [process, right of=agent1,xshift=2.75cm] {RL Agent 2 $\pi_2(o_2)$};
    \node (action_product) [io, below of=agent1,xshift=1.75cm,yshift=0cm] {$a=a_1 \times a_2$};
    \node (env) [process, below of=action_product,yshift=-0cm] {Environment};

    \node (obs) [io, below of=env,yshift=0cm] {$s,rew$};
    \node (obs1) [io,left of=obs,xshift=-3cm] {$o_1$, $rew$};
    \node (obs2) [io,left of=obs,xshift=5.2cm] {$o_2$,  $rew$};

    \draw [arrow] (agent1) |- node[anchor=east] {$a_1$} (action_product);
    \draw [arrow] (agent2) |- node[anchor=west] {$a_2$} (action_product);
    \draw [arrow] (action_product) -- node[anchor=west] {$a$} (env);
    \draw [arrow] (env) -- node[anchor=south] {} (obs);
    \draw [arrow] (obs) -- node[anchor=south] {} (obs1);
    \draw [arrow] (obs1) |- node[anchor=south] {} (agent1);
    \draw [arrow] (obs) -- node[anchor=south] {$o_1 = \mathbb{O}_1(s)$} (obs1);
    \draw [arrow] (obs) -- node[anchor=north] {$rew$} (obs1);
    \draw [arrow] (obs) -- node[anchor=south] {} (obs2);
    \draw [arrow] (obs) -- node[anchor=south]  {$o_2 = \mathbb{O}_2(s)$} (obs2);
    \draw [arrow] (obs) -- node[anchor=north]  {$rew$} (obs2);
    \draw [arrow] (obs2) |- node[anchor=south] {} (agent2);
    \end{tikzpicture}
 
    \caption{Parallel MARL setup with two agents performing a joint action $a=a_1\times a_2$, leading to a new state $s$ and reward $rew$, divided into observations $o_1$ and $o_2$.}
    \label{fig:cmarl}
\end{figure}

\begin{figure}[th]
\centering

    \begin{tikzpicture}[]
     {};
    \node (env) [process, below of=agent1,yshift=0.25cm,xshift=2cm] {Environment};
    \node (agent1) [process] {RL Agent 1 $\pi_1(o)$};
    \node (agent2) [process, below of=env,yshift=0.25cm,xshift=2cm] {RL Agent 2 $\pi_2(o)$};
    
    \draw [arrow] (agent1) -| node[anchor=west] {$a_1$} (env);
    \draw [arrow] (env) -| node[anchor=east] {$o_1 = \mathbb{O}(s_1),rew_1$} (agent1);

    \draw [arrow] (agent2) |- node[anchor=west] {$a_2$} (env);
    \draw [arrow] (env) |- node[anchor=east] {$o_2 = \mathbb{O}(s_2),rew_2$} (agent2);
    \end{tikzpicture}

\caption{Turn-based MARL system with two agents interacting sequentially with a shared environment. Each agent receives observations ($o_1$ and $o_2$) and rewards ($rew_1$ and $rew_2$) after their actions ($a_1$ and $a_2$).}
\label{fig:tmarl}
\end{figure}
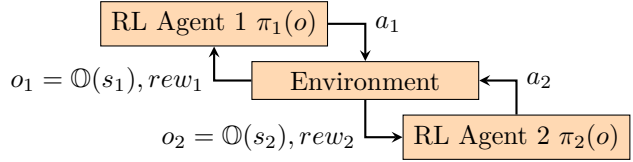

Generally, RL learns a near-optimal policy to achieve a fixed objective by acting and receiving rewards and observations from an environment simulator (see Example~\ref{ex:1} and Figure~\ref{fig:rl})~\citep{mnih2013playing}.
The RL setting can include a single agent~\citep{DBLP:journals/spm/ArulkumaranDBB17} or multiple agents (MARL)~\citep{DBLP:journals/aamas/ZhuDW24} that interact with the environment in parallel (see Figure~\ref{fig:cmarl}) or in a turn-based manner (see Figure~\ref{fig:tmarl}).

\begin{example}[RL Spaceship Agent]\label{ex:1}
    Let's imagine an RL agent learning to land a spaceship on the surface of a planet. 
    The agent gets information about position, angle, angular velocity, and whether the landing legs touch the ground.
    It can fire the main engine, left thruster, right thruster, or do nothing.
    During training, the agent earns rewards for landing in the target location.
\end{example}

Unfortunately, trained policies can exhibit \emph{unsafe behavior}~\citep{DBLP:conf/setta/GrossJJP22} like collisions (see Example~\ref{ex:2}), are vulnerable to \emph{attacks}~\citep{DBLP:journals/corr/HuangPGDA17} like noised observations from a defect sensor or attacker (see Example~\ref{ex:3}), and testing RL policies for the two previous mentioned issues consumes considerable effort in human and machine time~\citep{DBLP:conf/icst/AlshahwanHM23,DBLP:journals/corr/abs-2308-16557,DBLP:journals/corr/abs-2201-05371}.

\begin{example}[Unsafe Behavior]\label{ex:2}
A trained RL spaceship agent might exhibit unsafe behavior during landing, such as colliding with the surface at high speeds, since its reward function may focus too much on fast landing. Even though the agent might have learned a policy that achieves successful landings in ideal conditions, slight deviations in environmental parameters (e.g., terrain slopes or varying gravity) could cause the agent to perform risky maneuvers, like excessive lateral movement, resulting in collisions. \end{example}

\begin{example}[Adversarial Attack]\label{ex:3}
If an attacker introduces noise into the spaceship's observation system by tampering with the sensor data, the agent might receive inaccurate information about its position or angle. This can cause the agent to misjudge its actions, leading to unsafe decisions such as firing the thrusters in the wrong direction, which could result in a crash.
In this work,  we argue that adversarial perturbations—despite their structured nature—highlight the same type of vulnerability as random noise, namely, the model's sensitivity to particular inputs.
So, we assume that random noise is a special type of adversarial attack~\citep{DBLP:journals/access/OhashiNSYI21}.
\end{example}

\subsection{RL Policy Testing}
Fortunately, various studies have developed or adapted existing \emph{testing methods} to check RL policies using \emph{test cases} (see Example~\ref{ex:4}), which are specifically designed scenarios in which \emph{test oracles} determine the correctness of the outcomes~\citep{DBLP:conf/ast/MazouniSGA24,DBLP:conf/issta/PangYW22,DBLP:journals/tosem/BiagiolaT24,DBLP:conf/ecai/ZhouL23,DBLP:conf/issta/EniserGW0C22}.
Such oracles have diverse requirements, ranging from the observations or actions collected during the test cases to accessing the agent's internal states under test.

\begin{example}[Test Cases and Test Oracles in RL Testing]\label{ex:4} In testing an RL spaceship agent, a test case might involve a specific scenario where the spaceship starts at a high altitude with limited fuel. Given these challenging conditions, this scenario evaluates whether the agent can successfully land the spacecraft.
In this case, the test oracle could check if the spaceship successfully lands without crashing or running out of~fuel.
\end{example}

\emph{Search-based software testing (SBST)} leverages optimization techniques to automatically generate test cases that maximize the likelihood of uncovering faults in software systems~\cite{mcminn_stvr04}.
For instance, \citet{DBLP:journals/tosem/BiagiolaT24} propose a SBST method to find initial scenarios for which the policies fail to solve the environment (see Example~\ref{ex:5}).

\begin{example}[Search-Based Software Testing (SBST)]\label{ex:5} In the context of the RL spaceship agent, SBST can be used to identify initial conditions where the agent fails to land safely. For instance, SBST might generate scenarios where the spaceship starts at an unusually high altitude or with minimal fuel. By systematically exploring different initial conditions, SBST can find cases where the agent’s policy leads to unsafe behaviors, such as running out of fuel mid-flight or failing to stabilize its descent, helping developers identify weaknesses in the policy. \end{example}

\emph{Metamorphic testing (MT)} is geared towards situations where no oracle is available or is difficult to compute.
This technique relies on metamorphic relations to check how a program's input changes should affect the output~\citep{Chen1998,Chen2018,DBLP:conf/issta/EniserGW0C22} (see Example~\ref{ex:6}).
\begin{example}[Metamorphic Testing (MT)]\label{ex:6} In the RL spaceship agent scenario, MT could involve systematically altering the environmental conditions, such as doubling the gravity or adding wind forces, and observing how these changes affect the agent's ability to land. For instance, if the gravity is increased, the expected outcome is that the agent should use more thrust to counteract the increased downward force. MT ensures that the agent's policy adapts appropriately to these controlled environmental changes and highlights cases where the agent fails to adjust its actions as expected. \end{example}

\emph{Adversarial testing (AT)} aims to fool the agents and make them change their behaviors~\citep{DBLP:conf/ccs/ChanWY20}. 
In the context of RL policy testing, this is typically done by introducing so-called adversarial observations (see Example~\ref{ex:7}).
\begin{example}[Adversarial Testing (AT)]\label{ex:7} In the case of the RL spaceship agent, AT could involve introducing noisy sensor data during landing. For example, an adversarial observation might falsely indicate that the spaceship is at a higher altitude than it is, leading the agent to delay firing its thrusters. As a result, the agent could crash into the surface. Adversarial testing focuses on finding such vulnerabilities in the agent's policy by providing misleading observations and testing the robustness of its decision-making. \end{example}

\emph{Automated testing} approaches also exist to accelerate testing~\citep{DBLP:journals/tse/WangHCLWW24,DBLP:conf/icst/PaduraruPS21,DBLP:journals/corr/abs-2206-01335}, see Example~\ref{ex:9}.
While these methods reduce manual effort, using \emph{High-Performance Computing (HPC)}, which leverages powerful computational resources to handle large-scale tasks efficiently (see Example~\ref{ex:HPC}), can further accelerate testing~\citep{DBLP:journals/sigops/CiorteaZBCC09}.

\begin{example}[Automated Testing]\label{ex:9} Following the RL spaceship agent scenario, automated testing can accelerate testing the agent’s performance across various conditions. For example, an automated testing framework could create testing code and run thousands of simulated landings with different initial positions, fuel levels, and environmental factors like wind or gravity without requiring manual intervention. The system can automatically set up and track which scenarios lead to crashes or unsafe landings, flagging them for further analysis or automatically doing the analysis. This reduces the need for human oversight and speeds up the identification of edge cases and failure points in the agent’s policy. \end{example}

\begin{example}[HPC for RL Policy Testing]\label{ex:HPC}
Testing an RL spaceship agent across thousands of scenarios, such as varying gravitational forces or wind speeds, is computationally intensive. HPC accelerates this process by leveraging powerful computational clusters to execute simulations faster. For instance, HPC can simulate large amounts of landings in a fraction of the time required on standard hardware, enabling rapid identification of policy weaknesses and edge cases. \end{example}

In addition to the different testing categories, RL practitioners often need additional information about the test executions that would be collected during runtime.
For instance, logging RL policy executions enables detailed analysis, aiding iterative improvements by capturing actions, measuring performance, and detecting anomalies~\citep{DBLP:conf/issta/PangYW22}  (see Example~\ref{ex:8}).

\begin{example}[Logging RL Policy Executions]\label{ex:8} In the RL spaceship agent scenario, logging the agent’s policy executions allows developers to capture detailed information about each landing attempt. For instance, logs would record the agent's actions, such as firing the thrusters or adjusting the angle, along with the resulting changes in position and velocity. By analyzing these logs, 
developers gain better insight into the reasons for the failures, such as finding a correlation between failure occurrence and some action patterns.
This process would eventually improve the policy and make it more reliable.
\end{example}

\subsection{RL Policy Testing Problems}
Despite RL policy testing method advances~\citep{DBLP:journals/corr/abs-2312-09680}, there is no mature and unified tool support for various testing method categories like SBST, MT, and AT, logging RL policy executions, and allowing automated testing.

The primary reason is the broad diversity of the requirements of both the testing methods and the testing tasks~\citep{DBLP:journals/corr/abs-2312-09680} and differences between RL frameworks (compare, for instance, \citealt{towers_gymnasium_2023} with~\citealt{DBLP:conf/nips/TerryBGJHSSDHPW21}) (see also Example~\ref{ex:10}).

\begin{example}[Differences in Testing Methods for Single-Agent and Multi-Agent RL]\label{ex:10}
    Testing RL policies can differ substantially across frameworks like OpenAI Gym, Gymnasium, and others. For example, in OpenAI Gym, the environment step typically returns the observation, reward, done flag, and an info dictionary, allowing straightforward validation of policy performance. However, in Gymnasium, the step function distinguishes between terminated and truncated episodes.
    Additionally, the goals and methods for testing RL policies using MT, SBST, and AT vary between single-agent and multi-agent RL environments.
\end{example}

\subsection{Approach}
We developed a \emph{comprehensive testing framework for single-agent and multi-agent RL policies, covering various testing categories}. As part of this, we implemented \emph{Gimitest}\footnote{Repository: \url{https://github.com/DennisGross/Gimitest}}, an open-source tool that facilitates testing RL policies in single- and multi-agent scenarios using user-defined methods such as SBST, MT, or AT. 
Furthermore, it enables convenient logging execution and offers an interface for automated test code generation and testing approaches.
Its flexibility enables the utilization of HPC resources to efficiently execute large-scale simulations and computationally intensive testing tasks. This capability accelerates the analysis and validation of RL policies in both single- and multi-agent settings, leveraging systems like the Norwegian Innovation Cluster~\citep{cai2021ex3}.

\emph{Gimitest} takes trained policies, an environment, and a user-specified testing method as inputs, generating test results to identify faults (see Figure~\ref{fig:workflow}).
The tool assumes single-agent and multi-agent environment simulators to have \emph{action processing (step method)} and \emph{environment reset (reset method)} capabilities~\citep{DBLP:journals/corr/BrockmanCPSSTZ16,DBLP:conf/nips/TerryBGJHSSDHPW21}.
These assumptions align with the commonly used API in decision-making simulators (such as Farama Gymnasium).
The step method is called every time the RL policies interact with the environment, and reset is called every time the environment gets reset.
Gimitest enhances these simulator capabilities by decorating~\citep{DBLP:conf/sac/CharalampidouAA17} them with instances of the so-called ``GTest'' class.

\emph{Users can override the ``GTest'' class methods and apply modifications to environments, enabling them to create tests using their testing methods.}
An additional logger functionality allows test storage for further policy analysis, and a unified interface allows interaction with other systems, such as GPT-4~\citep{DBLP:journals/corr/abs-2303-08774} for automated test code generation and testing~\citep{DBLP:journals/tse/SchaferNET24} to save time.

\subsection{Applications}
\emph{Gimitest} is designed to support researchers and developers in testing single-agent and multi-agent RL systems.
It has already demonstrated its value in a replicability study~\citep{10.1145/3650212.3680382}, showing its potential to standardize and streamline the testing process.
The tool simplifies the implementation and adaptation of testing methods across a wide range of RL scenarios.

By enabling users to customize testing setups, automate repetitive processes, and integrate advanced analysis tools, Gimitest provides a framework for testing RL policies.
The following examples showcase how Gimitest can be applied to address different RL testing scenarios.

\begin{example}[Gimitest with SBST (similar for MT)]\label{ex:sbst}
Imagine testing an RL spaceship agent using Gimitest to identify failure conditions in the landing policy.
We applied SBST to discover the initial conditions where the spaceship agent fails to land safely.
The test involves generating scenarios with varying initial positions and fuel levels for the spaceship.
Unfortunately, most environment APIs do not allow the definition of such initial conditions.
Using Gimitest, however, we can create a custom subclass of the \emph{GTest} class and override the GTest method to configure the environment’s initial state at the beginning of each test run.
We could then conveniently explore various SBST methods, such as Random Testing, which generates random initial values for the spaceship's altitude and fuel.
By decorating the instantiation of the subclass to the used RL environment simulator, Gimitest automatically handles the interactions between the policy and environment, resetting the conditions after each test run.
Besides, as mentioned above, Gimitest unlocks systematic logging capabilities of any kind (after each run, throughout their executions, etc.), thanks to the additional policy and/or environment access functionalities made available by our tool. %
In short, Gimitest lets the practitioners conveniently apply their testing methods to RL agents.
\end{example}

The previous Example~\ref{ex:sbst} can be modified for MT in a straightforward manner.
Example~\ref{ex:12} illustrates how Gimitest facilitates AT by introducing noisy observations to test the resilience of RL policies against adversarial attacks.

\begin{example}[Gimitest with AT]\label{ex:12}
Imagine testing the robustness of an RL spaceship agent by introducing adversarial inputs designed to disrupt its decision-making.
Typically, we would apply AT in the test case execution of all environments \emph{manually}.
Precisely, we would simulate sensor noise or corrupted observations that could cause the agent to make unsafe landing decisions.
Using Gimitest though, we can create a custom subclass of the GTest-class to automatically inject the aforementioned adversarial noise into the spaceship’s sensor data during the environment’s step function.
For example, we would introduce slight distortions to the agent’s altitude readings, making the agent believe it's higher than it actually is.
By decorating the instantiation of the subclass to the used RL environment simulator, Gimitest executes the tests and logs scenarios where the adversarial noise causes the agent to make incorrect decisions, such as delaying the thruster firing and resulting in a crash.
\end{example}

The following example illustrates how Gimitest, with advanced AI systems like GPT-4, automates the testing process by analyzing environment parameters and generating test cases for rapid policy evaluation.

\begin{example}[Using Gimitest for Automated Testing with GPT-4]\label{ex:13} Gimitest integrates an automated testing pipeline, shown in Figure~\ref{fig:automated_testing}. The process begins with GPT-4 analyzing the environment's source code. For example, in the Cartpole environment, GPT-4 automatically extracts key parameters, such as gravity, cart mass, and pole length (with the help of Gimitest's functions). Once these parameters are identified, GPT-4 generates various test cases by systematically modifying these values to explore how the agent behaves under different conditions.
Gimitest executes these test cases automatically, running the RL agent in scenarios where the gravity is increased or the cart’s mass is altered. The pipeline generates the test code and executes the tests and logs the outcomes, providing insights into the agent’s robustness across a wide range of conditions.
\end{example}

Together, these examples highlight Gimitest's comprehensive approach to RL testing. From facilitating manual testing enhancements to enabling full automation, Gimitest supports researchers and practitioners in RL testing across various scenarios.

\subsection{Outlook}
The remainder of the paper is structured as follows: first, we introduce the background necessary to understand the paper, followed by a survey of related work. Next, we present the functionality and architecture of Gimitest. We then demonstrate the effectiveness of our proposed testing framework and tool through multiple use cases. Finally, we conclude the paper and provide an outlook for future work.

\section{Background}
In this section we introduce the fundamentals for this paper.

\subsection{Reinforcement Learning (RL)}
For the context of our work, reinforcement learning (RL) addresses sequential decision-making problems that can be modeled as a Markov decision process (MDP)~\citep{sutton2018reinforcement}. The RL agent $\pi$ interacts with the environment, which represents the sequential decision-making problem, via a series of interactions, i.e., steps, where it first observes the current state, then issues an action, and then receives a reward and the observation of the next state (see Figure~\ref{fig:rl}).
The MDP is represented by a four-tuple $(\mathcal{S}, \mathcal{A}, R, P)$, where $\mathcal{S}$ is the environment's state space, $\mathcal{A}$ is the action space of all possible actions $a$ the agent can take, $R: \mathcal{S}\times\mathcal{A}\times\mathcal{S}\to\mathbb{R}$ is the reward function, assigning a numeric reward to an action taken in given current and next states, and $P: \mathcal{S}\times\mathcal{A}\to\mathcal{S}$ is the potentially stochastic, transition function, that determines the next state when taking action in the current state.
Additionally, we consider the observation function $\mathbb{O}: \mathcal{S}\to\mathcal{O}$ that maps the fully-informed state to a (partial) observation, which might not include all information about the state. For example, the hidden dynamics of a physical system might be part of the state, but not of the RL agent's observation.
The goal of a RL agent is to maximize the accumulated rewards over an infinite horizon.

Following the above definition, it can be extended to the parallel and turn-based Multi-Agent Reinforcement Learning (MARL) settings, as indicated in Figures~\ref{fig:cmarl} and \ref{fig:tmarl}.

\subsection{Software Testing Techniques}
Various testing techniques are available to ensure the reliability and safety of RL policies.
Common methods include Search-Based Software Testing (SBST), Metamorphic Testing (MT), and Adversarial Testing (AT), which exposes weaknesses by simulating malicious or unexpected inputs.

\subsubsection{Search-based Software Testing (SBST)}
When testing complex software systems that cannot easily be analyzed, it can be relevant to guide the search of relevant test cases with program executions and fitness function evaluations \citep{mcminn_stvr04}. 
In the case of RL policies, the purpose of search-based software testing is to find input tests, such as environment states, configurations or any kind of data representing a particular situation from which the agent under test fails to solve the decision-making task~\citep{zolf23,DBLP:conf/issta/EniserGW0C22,DBLP:conf/ast/MazouniSGA24}. 
Relying on ML models and genetic algorithms, SBST approaches can guide the search toward faulty sequences of states and actions produced by the RL agent.

\subsubsection{Metamorphic Testing (MT)}

In cases where the expected outcome of the system-under-test, i.e., in our context the RL policy, cannot be exactly specified, i.e., there is no \textit{oracle}, we encounter the \textit{oracle problem} in software testing~\citep{Weyuker82,Barr2015}.
One approach to overcome the oracle problem is MT, where, instead of specifying the exact oracle of a test case, we specify metamorphic relations and generate \emph{follow-up test cases} by using these relations \citep{Chen1998,Chen2018}. 
\begin{definition}[Metamorphic Relation (MR)]
Let $P$ be the program under test, and $x$ and $y$ two test inputs for $P$. A MR for $P$ is then expressed as a relation $\forall x, \forall y, r_i(x,y) \implies r_o(P(x), P(y))$, where $P(x)$ respectively $P(y)$ denotes the execution of $P$ on $x$ respectively $y$. 
\end{definition}
Keep in mind that MRs are necessary, but not sufficient, properties to ensure the correctness of $P$ w.r.t. its specification, i.e., $r_i(x,y) \wedge \neg r_o(P(x),P(y)) \implies \neg correct(P)$. Nevertheless, MRs are convenient properties for generating follow-up test cases. 
\begin{definition}[Follow-up test cases]
Let $r_i(x,y) \implies r_o(P(x), P(y))$ be a MR for $P$. If there exists a transformation $f$ (possibly non-deterministic) such that $y=f(x)$, then it becomes possible to generate a sequence of {\it follow-up test cases} from $x$, namely $<x,t(x), t(t(x)), ...>$ which all have to fulfill the MR for $P$. Thus, the transformation $t$ is convenient for generating follow-up test cases.
\end{definition}

The effectiveness of MT depends on the strength of the MRs.
While a lot of MRs can be found for a given system~\citep{Segura2016}, many of them tend to be ineffective in revealing fault, albeit that they focus on trivial properties or have a too wide criterion to identify violations.
Still, MT has been effective for the testing of many machine learning applications, ranging from simple classifiers~\citep{Murphy2008,Xie2011}, deep learning models in general~\citep{Ding2017a}, machine translation \citep{Sun2018}, or computer vision tasks \citep{Spieker2020,Spieker2024}.
MT has also been successfully applied for policy testing~\citep{DBLP:conf/issta/EniserGW0C22,Eisenhut2023AutomaticMT,DBLP:journals/corr/abs-2312-09680}, giving a strong motivation for its inclusion in our framework.

\subsubsection{Adversarial Testing (AT)}

Informally, in Machine Learning \emph{adversaries} describe inputs that fool the learned models under attack.
The vulnerability of Neural Networks (NNs) to adversarial inputs was first observed by~\citet{szegedy2014intriguingpropertiesneuralnetworks}, where they found that unperceivable random perturbations in images can lead NN classifiers to misclassify them. 
\citet{goodfellow2015explaining} further extend this work by suggesting additional causes of this vulnerability, as well as introducing fast methods for generating adversarial examples.
In particular, they developed the fast gradient sign method (FGSM), which consists of perturbing the inputs up to a maximum $\epsilon$, signed by the gradient of the loss between the output of the models for the initial inputs and the ground truth.
The intuition is that such perturbations maximize the change in the model's output.
This method was then adapted to the reinforcement learning setting by~\citet{DBLP:journals/corr/HuangPGDA17}.
Here, the idea is to bypass the limitation of not having observations labeled with the optimal action distribution by assuming that the attacked policy is optimal.
Therefore, assuming stochastic policies $\pi: \mathcal{O}\to\mathcal{A}$, the loss is computed between the output $\pi(o)$ and a distribution that places all weights on the highest-weighted action in $\pi(o)$.
Consequently, in this setting, adversarial testing aims to undermine the performance of the RL policy, such as reducing its episodic accumulated rewards.

\section{Related Work}\label{sec:related}
In the following, we first cover RL policy testing methods (that Gimitest would ease the implementation and application) before detailing tools related to Gimitest.

Single-agent testing has been largely covered by the research community.
\emph{DeepTest}~\citep{DBLP:conf/icse/TianPJR18} is an automated testing framework of Deep Neural Networks (DNNs) for autonomous driving cars based on
Metamorphic Testing.
It specifically targets models that input images.
DeepXplore also tests image input-based DNN agents but relies on differential testing to find errors (instead of MT)~\citep{10.1145/3361566}.
Therefore, DeepXplore inputs several policies and finds failures by comparing their outputs.

MDPFuzz is a black box fuzz testing framework for policies tackling MDPs, verifying if the target policy reaches abnormal or dangerous states~\citep{DBLP:conf/issta/PangYW22,10.1145/3650212.3680382}.
It features a cover model of the state sequences observed during testing to guide the fuzzing process towards unseen behaviors.
Similarly to MDPFuzz, \citet{DBLP:conf/issta/EniserGW0C22} and \citet{Eisenhut2023AutomaticMT} use fuzzing techniques to generate test inputs, whose results are however then checked with MT.

Recently, \citet{DBLP:conf/ast/MazouniSGA24} and \citet{bhatt2022deepsurrogateassistedgeneration} explore the use Quality-Diversity optimization~\citep{10.3389/frobt.2016.00040} to search for failures.
On the other hand, RL itself has been used to optimize the search.
For instance, DeepCollision trains a RL agent to modify the environment during the execution of the test cases to find failures~\citep{9712397}.
Likewise, \citet{10.1109/ICSE48619.2023.00155} propose MORLOT, a framework for handling many-objective policy testing tasks. 
Their approach consists of training a Deep Q-Learning agent~\citep{mnih2013playing} to find failure-triggering configurations of the environment.

Other types of SBST approaches include genetic algorithm~\citep{zolf23, DBLP:journals/tosem/BiagiolaT24}, depth-first search~\citep{tappler2022searchbasedtestingreinforcementlearning} and test oracle approximation through AI planning heuristics~\citep{DBLP:conf/aips/SteinmetzFEFGHH22}.
We refer interested readers to the survey of~\cite{DBLP:journals/corr/abs-2312-09680} for a broader review of policy testing methods.

In the multi-agent setting (MARL), \emph{Melting Pot} tests learned policies' generalization to unseen situations~\citep{DBLP:conf/icml/LeiboDVASKMBMG21}.
\citet{DBLP:conf/ecai/ZhouL23} present RTCA,  a method that tests if the trained agents are robust against noise state observations.
RTCA introduces two key innovations: 1) a Differential Evolution (DE) method to select critical agents as noise victims and determine worst-case joint actions; 2) a team cooperation policy evaluation used as the DE optimization objective. Adversarial state perturbations are then applied to the critical agents based on these joint actions.

Adversarial testing has also been extensively applied to RL algorithms.
The first line of research aims to reduce the number of times adversaries are injected (while achieving similar results compared to systemic attacks).
For instance, \citet{kos2017delvingadversarialattacksdeep} refine the FGSM method by only attacking critical states during the episodes, which they select based on the value function of the policy.
On the other hand, \citet{lin2019tacticsadversarialattackdeep} identifies when to attack the policy by targeting states that maximize the difference between the most preferred action and the least preferred action.
\citet{sun2020stealthyefficientadversarialattacks} opt for another strategy for planning the attacks and emphasize long-term damage impacts.
They propose to achieve this goal by learning an antagonist model, which they train with the reward function to account for the agent's end goal.
Another line of work aims at the adversaries themselves.
For instance, \citet{10.1145/3320269.3384715} quantify how much changing each feature composing the observations affects the rewards with a \emph{static reward impact map} to only perturb the most impactful features.
Similar to testing, there exists also works that formally verify adversarial attacks on RL policies in both single-agent~\citep{DBLP:conf/icaart/GrossS0023} and multi-agent settings~\citep{DBLP:conf/aips/GrossS0023}.

Every framework has its own specifications for implementing the initialization and execution of the test cases.
Gimitest offers a unified framework that is suitable for all of them since it builds on the RL community's common interface. %

Related to our work, \emph{COOL-MC} enables the training of single-agent and multi-agent RL policies in modeled environments while formally verifying them at any stage of the training process~\citep{DBLP:conf/setta/GrossJJP22}.
\emph{MoGym} is an integrated toolbox that trains and verifies RL agents using deep statistical model checking~\citep{DBLP:conf/tacas/KohlKH21}.
\emph{Mungojerrie} is a tool used for testing reward functions, specifically for $\omega$-regular objectives~\citep{DBLP:journals/corr/abs-2106-09161}.
\cite{DBLP:journals/sttt/GuJPSEL22} introduce MCRL, merging verification with MARL for mission planning to ensure safety-critical requirements.
Our tool differs in that it tests RL policies rather than verifying them for specific behaviors.

\section{Functionality}\label{sec:func}
\begin{figure}[tbp]
    \begin{center}

\begin{tikzpicture}[]

\node (pro2) [process2,xshift=0.9cm,yshift=0.2cm] {Train Policies};
\node (pro3) [process2, right of=pro2,xshift=2cm] {Test Policies};

\node (in2) [io,align=center,above of=pro2,yshift=0.6cm] {Environment};
\node (in3) [io,align=center,above of=pro3,yshift=0.6cm] {Testing Method};
\node (dec1) [decision, right of=pro3,xshift=2cm] {Faulty?};
\node (stop) [startstop, right of=dec1,xshift=2.5cm] {Deploy Policies};

\draw [arrow] (in2) -- node[] {} (pro2);
\draw [arrow] (in3) -- node[] {} (pro3);
\draw [arrow] (pro2) -- node[] {} (pro3);
\draw [arrow] (pro3) -- node[] {} (dec1);
\draw [arrow] (dec1) -- node[anchor=south] {no} (stop);
\draw [arrow] (dec1.south) -| node[anchor=north]{yes} (pro2.south);

\draw [dashed, very thick, red] ([shift={(-1.1,-0.75)}]in3.north west) rectangle ([shift={(0.65,-1)}]dec1.south east);
\node[text=red] at (3.3, -1.15) {\emph{Gimitest}};

\end{tikzpicture}

    \end{center}
    \caption{Gimitest's role within the RL workflow, which involves training, testing, and deploying policies. The flow starts with training policies using an environment (not part of Gimitest), then testing the policies through a user-specified testing method. Gimitest, highlighted in a red dashed box, is responsible for the testing phase, where it identifies potential faults. If the policies are deemed non-faulty, they proceed to deployment. Otherwise, the workflow loops back to policy retraining.}
    \label{fig:workflow}
\end{figure}
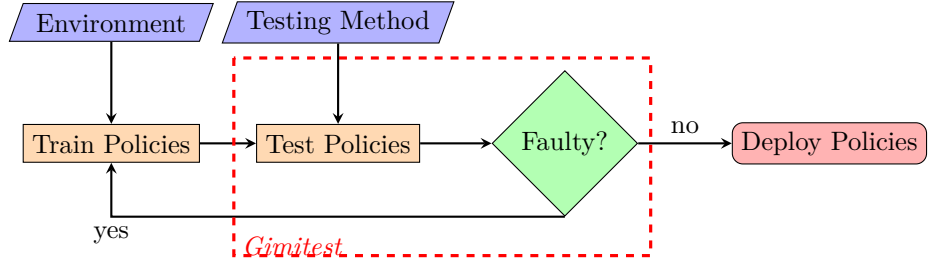

\emph{Gimitest} enables single-agent and multi-agent RL policy testing.
It takes trained policies, an environment, and a user-specified testing method as inputs to generate test results and identify potential faults (see Figure~\ref{fig:workflow}). 
Gimitest can be applied to commonly used APIs in decision-making simulators, i.e., the Farama Gymnasium and OpenAI Gym interface, requiring only access to the \emph{action processing (step method)} and the \emph{environment reset (reset method)}.
Technical details are provided in Section~\ref{sec:architecture}.

Based on the different testing methods (see also related work in Section~\ref{sec:related}) and general interaction between policies and environments (see Figure~\ref{fig:rl}, Figure~\ref{fig:cmarl}, and Figure~\ref{fig:tmarl}), we created a \emph{general testing framework} for single-agent and multi-agent RL policies.
This framework consists of the following steps:
\begin{enumerate}
    \item \emph{Pre-configure and test the environment state:} Setting up the (initial) state.
    \item \emph{Environment Execution:} Executing the environment step with the RL policy action.
    \item \emph{Post-configure and test the environment state:} Modifying potential observations, configuring the current environment state, and testing.
\end{enumerate}
Based on this framework, we support the following functionality.

\subsection{Search-based Testing (SBST)}
SBST aims to find initial scenarios for which the policies under test fail to solve the environment or exhibit unsafe behaviors~\citep{DBLP:journals/tosem/BiagiolaT24}.
It thus involves searching for the \textit{input tests} for initializing the environment so that, when evaluated with known test oracles, e.g., failure conditions, failures are revealed.
These inputs are environment configurations, defining the agent's initial environment state.
\emph{Gimitest allows the configuration and testing of the environment state at every time step in the RL policy execution.}
How the search is performed depends on the testing methods (see Section~\ref{sec:related}).
For an example, we refer to Example~\ref{ex:sbst}.

\subsection{Metamorphic Testing (MT)}
In MT, metamorphic relations describe how input changes should affect output, eliminating the need for a test oracle~\citep{DBLP:conf/issta/EniserGW0C22}.
The process involves running the RL policy in an original scenario, making systematic modifications to create follow-up cases, and observing behavior changes.
\emph{Gimitest enables MT by modifying and testing the environment state at each time step.}

\subsection{Adversarial Testing (AT)}
An \emph{attack} maps an observation to an \emph{adversarial observation}~\citep{DBLP:journals/corr/HuangPGDA17}. 
A successful attack at a given state leads to a \emph{misjudgment} of the RL policy, and an attack is bounded by constraints.
\emph{Gimitest lets the user access the internals of the policy for white-box attacks, as well as transform the policy's observation for black-box attacks at every time step.}
For an example, we refer to Example~\ref{ex:12}.

\subsection{Logging} Logging RL policy executions enables detailed analysis, aiding iterative improvements by capturing actions, measuring performance, and detecting anomalies~\citep{DBLP:conf/issta/PangYW22}. It provides insights into decision-making and environment interactions, facilitating effective debugging and tuning of RL models~\citep{DBLP:conf/aips/SteinmetzFEFGHH22}.
\emph{Gimitest enables the logging of RL executions.}

\subsection{Automated Test Code Generation}
Gimitest aims to streamline testing for RL developers by allowing automatic test code generation~\citep{DBLP:journals/tse/SchaferNET24} and testing.
\emph{It enables automatic retrieval of environment source code, Gimitest source code, and supports specific test method implementations, allowing the automated generation of RL policy tests.}
For an example, we refer to Example~\ref{ex:13}.

\section{Architecture}\label{sec:architecture}
The \emph{Gimitest} framework is written in \emph{Python 3} and enables testing RL policies in \emph{Farama Gymnasium}~\citep{towers_gymnasium_2023}, \emph{PettingZoo}~\citep{DBLP:conf/nips/TerryBGJHSSDHPW21}, and \emph{similar RL environment simulators}~\citep{DBLP:journals/corr/abs-1903-03176,DBLP:journals/corr/BrockmanCPSSTZ16}.

\begin{figure}[tbp]
\centering

    \begin{tikzpicture}[]
        \node (env) [process2, draw, minimum width=2.5cm, minimum height=0.5cm] {Environment};

        \node (leftcircle) [circle, draw, left of=env, xshift=-1.5cm] {Pre};
        \node (rightcircle) [circle, draw, below of=env,yshift=-0.1cm] {Post};

        \node (leftprocess) [process2, draw, left of=leftcircle, xshift=-2cm, minimum height=0.5cm] {Policies};

        \draw [arrow] (leftprocess.east) -- node[anchor=north]{} (leftcircle.west);
        \draw [arrow] (leftcircle.east) -- node[anchor=north]{} (env.west);
        \draw [arrow] (env.south) -- node[anchor=north]{} (rightcircle.north);
        \draw [arrow] (rightcircle.west) -| node[anchor=north]{} (leftprocess.south);
    \end{tikzpicture}

\caption{Interception of \emph{Gimitest} in the RL policy-environment interaction. \emph{Pre} refers to functionality added before the environment is called; \emph{post} refers to functionality added afterward.}
\label{fig:enhances}
\end{figure}
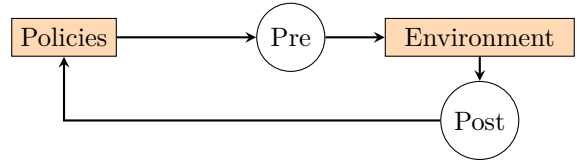

We implement the \emph{decorator design pattern}~\citep{DBLP:conf/sac/CharalampidouAA17} to extend the reset and step capabilities of single-agent and multi-agent environments (see Algorithm~\ref{fig:decorater_pattern}) with instances of the \emph{GTest class}, allowing for testing at specific time intervals.
This pattern dynamically enhances an object's functionality at runtime  (see Figure~\ref{fig:enhances}), promoting code reusability without altering its structure.

\begin{algorithm}
\caption{The decorator design pattern is illustrated in decorating the CartPole environment from Farama. The environment is decorated by GTest to enable testing, and GTest is further decorated by GLogger to allow the test results to be logged.}
\begin{algorithmic}[1]
\STATE env = gym.make('CartPole-v1')
\STATE m\_gtest = GTest() 
\STATE EnvDecorator.decorate(env, m\_gtest)
\STATE m\_logger = GLogger("m\_log")
\STATE GTestDecorator.decorate\_with\_logger(m\_gtest, m\_logger)
\end{algorithmic}
\label{fig:decorater_pattern}
\end{algorithm}

\begin{figure}
    \centering
    \includegraphics[width=0.75\textwidth]{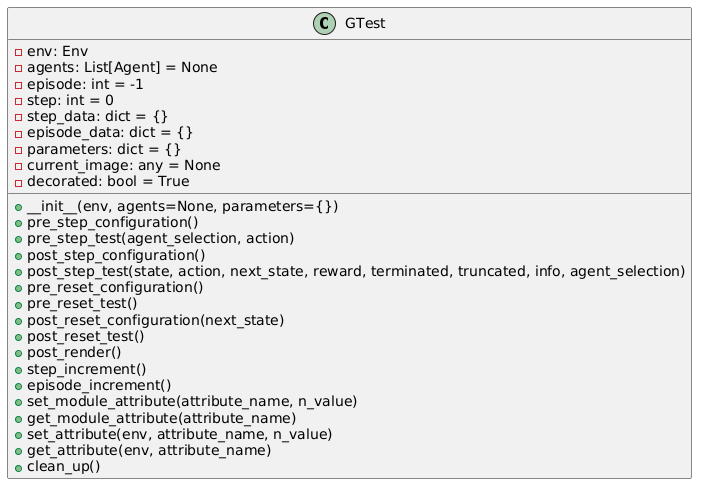}
    \caption{GTest base class.}
    \label{fig:gtest_class}
\end{figure}

GTest is a base class (see Figure~\ref{fig:gtest_class}).
A base class is a class in object-oriented programming from which other classes (called derived or child classes) can inherit properties and behavior~\citep{rentsch1982object}. The base class defines attributes and methods that are common to all its derived classes, allowing code reuse and reducing redundancy.

We can create custom GTest subclasses and override specific methods to suit our testing needs.
Overwriting the \emph{GTest configuration methods} allows us to modify the environment at specific time steps, and overwriting the \emph{GTest testing methods} allows us to test the policy at specific time steps.

By applying GTest decoration to the environment step method, the environment sequentially performs the pre-step-configuration, pre-step-test, original step method, post-step-test, and post-step-configuration GTest calls (see Figure~\ref{fig:enhances}).
Applying GTest decoration to the environment reset method, the environment sequentially executes the pre-reset-test, pre-reset-configuration, original reset method, post-reset-test, and post-reset-configuration GTest calls.

\begin{figure}
    \centering
    \includegraphics[width=0.8\textwidth]{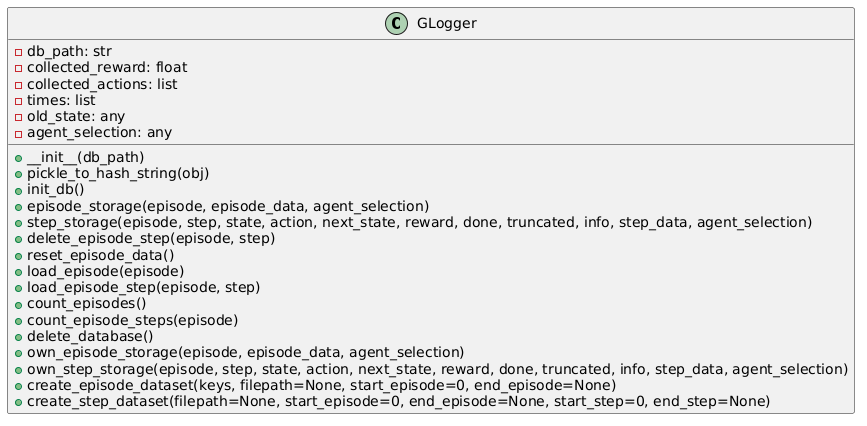}
    \caption{GLogger base class.}
    \label{fig:glogger_class}
\end{figure}
\emph{GLogger} allows us to log the whole testing process at every point in time by decorating GTest instances.
Users can override \emph{GLogger} methods (see Figure~\ref{fig:glogger_class}) to implement customized logging behaviors, such as filtering specific events at specific time steps or adding new log outputs tailored to the needs of specific tests via, for instance, at every time step by overriding own\_step\_storage().

While custom environments are supported, environments that rely on third-party simulators, such as the Carla simulator for autonomous driving~\citep{DBLP:conf/corl/DosovitskiyRCLK17}, have not been studied yet.
Parallel test execution is not supported yet. 

\section{Use Cases}
We showcase \emph{Gimitest's} features in single-agent and multi-agent settings, demonstrating its usability for various policy evaluation tasks and environments.
Specifically, we explore three single-agent testing scenarios: failure detection with three SBST methods, AT, and MT.
Additionally, we highlight \emph{Gimitest's} support for parallel MARL, turn-based MARL, and automated code generation.

We use the official Farama Gymnasium~\citep{towers_gymnasium_2023} and PettingZoo~\citep{DBLP:conf/nips/TerryBGJHSSDHPW21} environment versions.
For failure detection, we use the \emph{Lunar Lander} environment; for AT, we use \emph{Mountain Car}; for MT and automated testing, we use \emph{Cart Pole}; for turn-based MARL SBST, we use \emph{Connect Four}; and for parallel MARL SBST, we use \emph{Waterworld}.

We provide the trained policies in our repository with further details on how to run the experiments.

\subsection{Failure Detection via SBST}\label{subsec:sbst}
In \emph{Lunar Lander}, the task is to safely land a spacecraft on the Moon from a start position above a flat landing area.
To do so, the agent can fire the different engines of the spacecraft.
We test a well-trained policy made publicly available by the RL community~\citep{rl-zoo3}.
We reduce the possible initial scenarios by fixing the shape of the landscape and only varying the forces applied to the spacecraft within a range of $[-1000, 1000]^2$.
As for detecting failures, they occur if the lander crashes or is no longer visible.
We use three SBST methods to find fault-triggering inputs: an evolutionary search that starts with a random population and then selects those with the lowest rewards to identify weak runs (ST); random testing (RT), which continuously generates random inputs to test the agent; and a fuzz-based search (FT), which implements the previously mentioned MDPFuzz~\citep{DBLP:conf/issta/PangYW22} without coverage guidance.
Roughly, this method maintains a pool of inputs that reveal weaknesses and prioritizes highly sensitive ones.

\begin{figure}[th]{}
        \centering
        \includegraphics[width=0.5\textwidth]{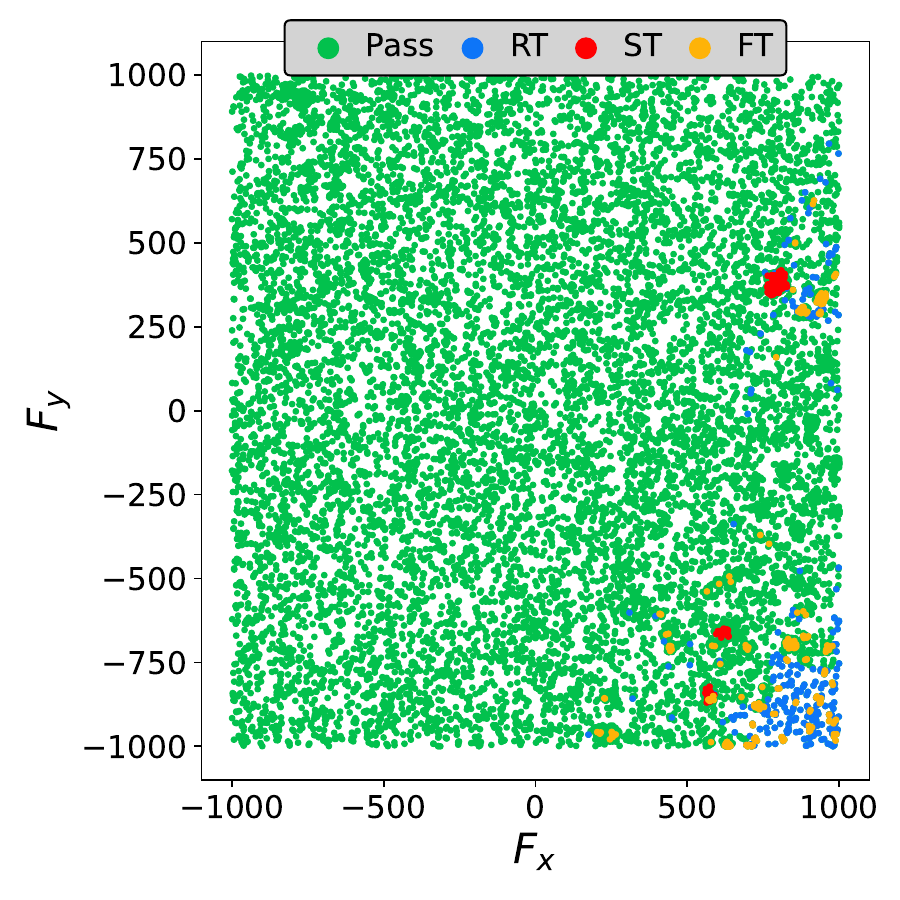}
        \caption{
        SBST: Distribution of the input tests for the Lunar Lander use case.
        The inputs describe the initial forces applied to the agent on the x and y axes.
        Passing inputs are shown in green, with fault-triggering ones colored blue, red, and orange for random ST, evolutionary ST, and FT, respectively.
        }
        \label{fig:sbt_faults}
    \end{figure}

Figure~\ref{fig:sbt_faults} summarizes the results.
It shows the fault distribution over the input space ($[-1000, 1000]^2$) as well as all the safe executions.
As expected, overall, the policy manages to safely land the spacecraft for most of the initial forces tested.
If we now compare the faults found by the three methods, we observe that they are distributed differently.
In particular, the evolutionary and fuzzing searches identify specific clusters of failures, while random testing reveals that the policy is especially unsafe when the spacecraft is pushed downwards to the right.
Note that the frameworks conveniently execute the test cases thanks to the automated calls of the post-reset-configuration function of the GTest class implemented for this use case.

\subsection{Adversarial Testing}
\begin{figure}[th]{}
        \centering
        \includegraphics[width=0.5\textwidth]{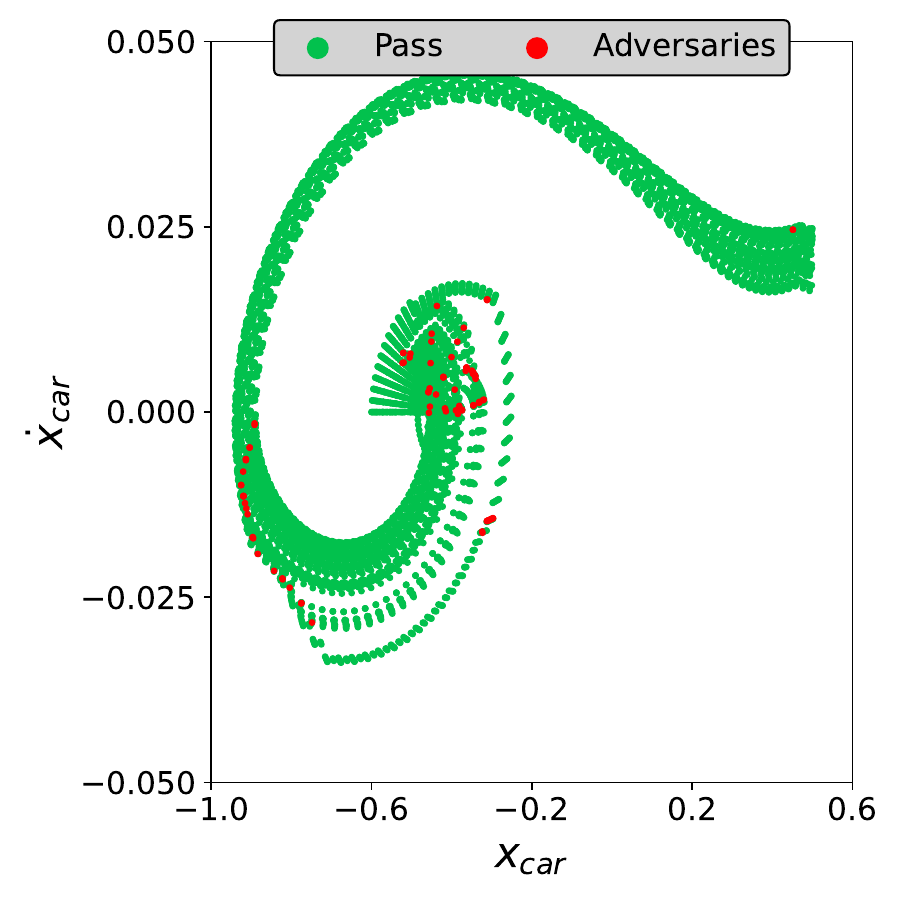}
        \caption{
        Results of the Adversarial Testing experiment on the Mountain Car policy.
        Adversarial states are shown in red, and the ones for which the policy does not change its behavior are shown in green.
        The states are projected in 2D \textit{w.r.t} the position of the car (horizontal axis) and its velocity (vertical axis) they describe.
        }
        \label{fig:adv_faults}
    \end{figure}

In the \emph{Mountain Car} environment, the agent's goal is to drive a car to the top of a hill from a valley.
The policy fails to solve the problem if it could not have the vehicle to the top after 200 time steps.
We attack the policy using the \emph{Fast Gradient Sign Method (FGSM)}~\citep{DBLP:journals/corr/HuangPGDA17}.
That is, the agent is automatically attacked after each action thanks to the post-action method made available by \emph{Gimitest}.
Figure~\ref{fig:adv_faults} shows the agent's safe and adversarial states regarding the position and velocity of the car.
We can see that most adversaries are detected at initial states (central red dots) and as the agent builds momentum to exit the valley (left red points).

\subsection{Metamorphic Testing}
We adopt the method from \citet{DBLP:conf/issta/EniserGW0C22} to identify bugs in a policy trained on the \emph{Cart Pole} environment.
In this approach, bugs are defined as simpler scenario versions causing failures. 
In other words, a test case $c$ reveals a bug in the policy if the latter solves $c$ but fails for the easier test case $c'$.
As such, this method requires defining un/relaxation operations specific to the environment.
In the \emph{Cart Pole} environment, where the goal is to balance a pole on a moving cart, we simplify states by reducing the pole's angle \textit{w.r.t} the vertical axis. 
Our testing uncovers over 100 such bugs, typically with higher or negative initial pole angles and positions.
While being outside the scope of this paper, these findings suggest that we could enhance the robustness of the agent through policy retraining or repair.

\subsection{Policy Analysis}
\begin{figure}[th]{}
        \centering
        \includegraphics[width=0.45\textwidth]{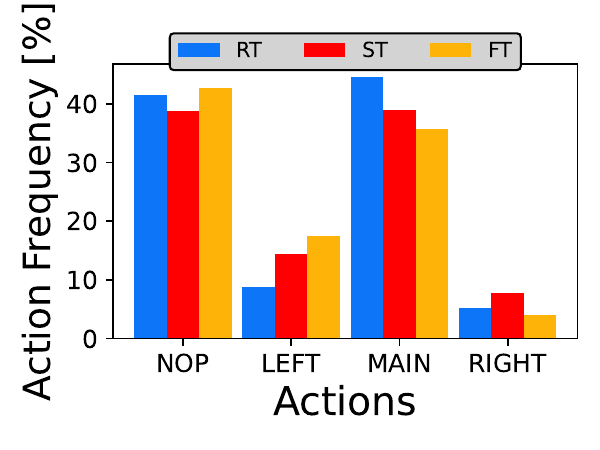}
        \caption{
         Action frequencies of the \emph{Lunar Lander} policy, when tested with random, evolutionary, and fuzz SBST methods.
         ``NOP'' refers to the do-nothing action, while ``LEFT'', ``MAIN'' and ``RIGHT'' refer to firing the three engines of the spacecraft.
        }
        \label{fig:action_distribution}
    \end{figure}

We exemplify the logging features of \emph{Gimitest} by automatically collecting the actions performed by the policy tested in the \emph{Lunar Lander} environment (see Subsection~\ref{subsec:sbst}).
Figure~\ref{fig:action_distribution} displays the action frequencies for the three SBST methods.
Automated via GTest's post-step-test function, data reveals that evolutionary-based and fuzz-based approaches prompt more 
frequently left engine firing than random testing, which instead favors the main engine.

\subsection{MARL Testing}
\emph{Connect Four} is a strategy game where two players take turns dropping colored discs into a vertical grid. The goal is to align four of your discs in a row, either horizontally, vertically, or diagonally, before your opponent.
\emph{Waterworld} is a simulation in which archaea navigate their environment, striving to survive. These archaea, known as pursuers, aim to consume food (evaders) while avoiding poison.
In this simulation, the pursuers are the agents, and both the food and poison are elements of the environment\citep{DBLP:conf/nips/TerryBGJHSSDHPW21}.

\begin{figure}[th]{}
        \centering
        \includegraphics[width=0.5\textwidth]{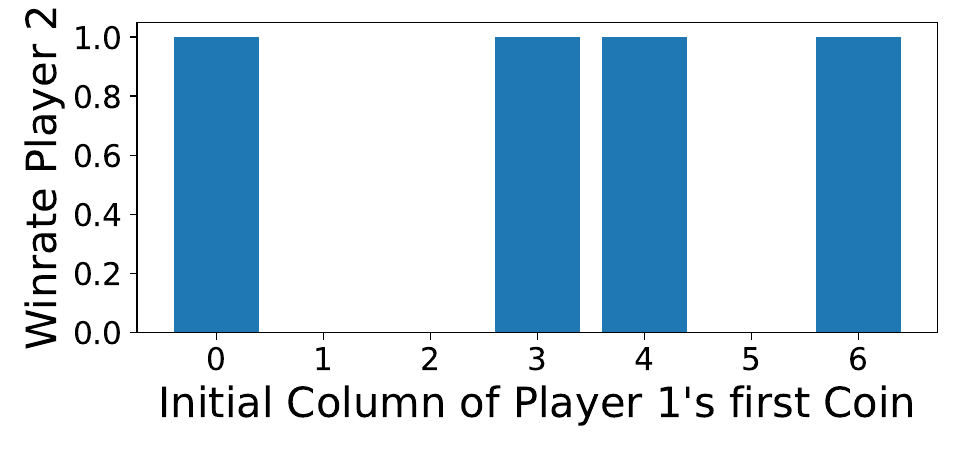}
        \caption{
        Connect Four results. The win rate for player 2 is on the y-axis, and the x-axis describes the different initial positions of player 1 coins.
        }
        \label{fig:tmarl_testing}
    \end{figure}

\begin{figure}[th]{}
        \centering
        \includegraphics[width=0.5\textwidth]{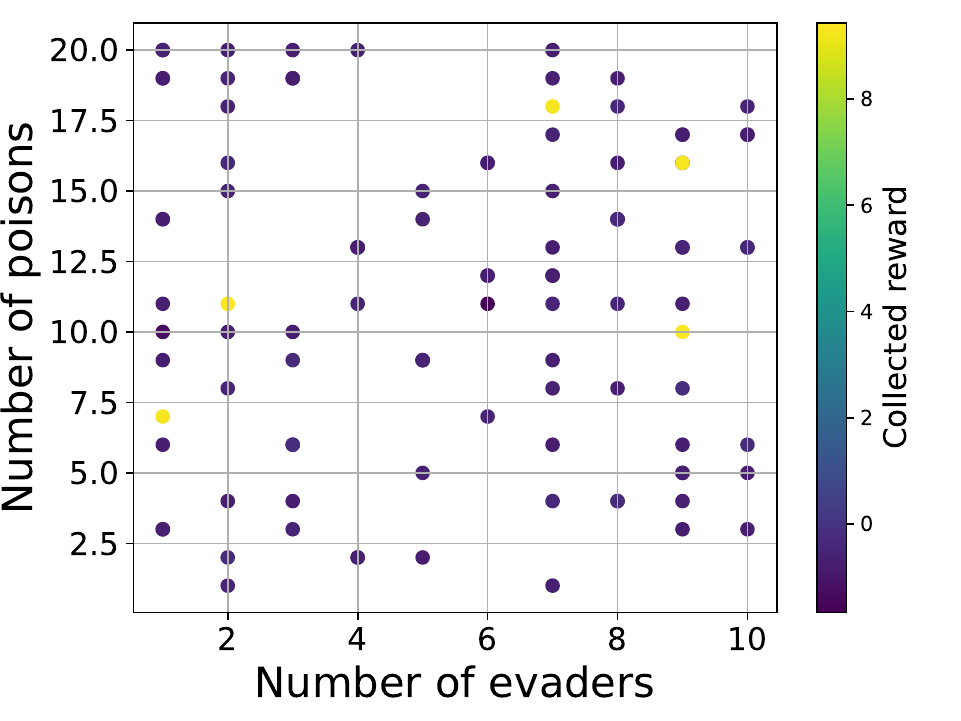}
        \caption{
        Waterworld results. The x-axis represents the different numbers of evaders, and the y-axis represents the different numbers of poisons in the environment. The color indicates the number of collected rewards for the given evader-poison setting.
        }
        \label{fig:pmarl_testing}
\end{figure}

We trained PPO policies in the turn-based MARL environment Connect Four and in the parallel environment Waterworld.
In Connect Four, we tested the success rate of player agent 2 for differently positioned starting coins of player agent 1 (see Figure~\ref{fig:tmarl_testing}). Our testing results show that with the current trained MARL policies, player 2 always wins when the coin is set into columns 0, 3, 4, and 6, and in the other cases, player 2 loses.
In Waterworld, we tested how well the trained MARL policies performed with different numbers of poisons and food sources (see Figure~\ref{fig:pmarl_testing}).
We observed, for instance, that the agents rarely collected more than 8 reward points.

\begin{figure}[th]{}
        \centering
        \includegraphics[width=0.5\textwidth]{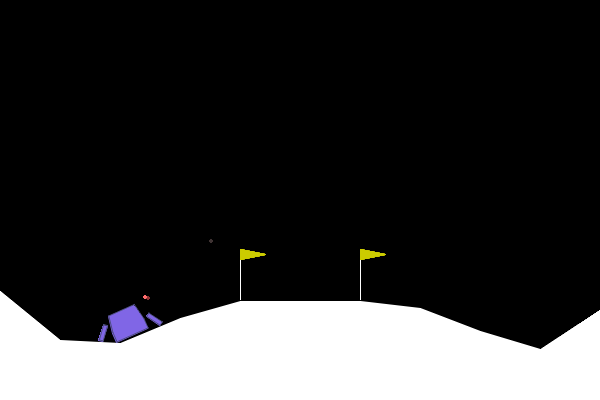}
        \caption{
        This recorded frame shows a failure. Analyzing this frame with GPT-4 helps categorize the failure, such as \emph{the spacecraft landed on the left side of the landing pad.}
        }
        \label{fig:failure_explaination}
    \end{figure}

\subsection{Automated Test Code Generation}
\begin{figure}[th]
    \begin{center}

\begin{tikzpicture}[]

\node (in1) [io,align=center] {Environment};
\node (extract) [process2, below of=in1,yshift=0cm] {Extract Parameters};
\node (generate) [process2,right of=extract, xshift=2.75cm] {Test Generation};

\node (execute) [process2, right of=generate, xshift=2.5cm] {Test Execution};
\node (in2) [io,align=center, above of=execute,yshift=0cm] {RL policy};

\node (stop1) [startstop, left of=execute, xshift=4.25cm] {Test Results};

\draw [arrow] (in1) -- node[anchor=west] {} (extract);
\draw [arrow] (in2) -- node[anchor=west] {} (execute);
\draw [arrow] (extract) -- node[] {} (generate);
\draw [arrow] (generate) -- node[midway,right] {} (execute);
\draw [arrow] (execute) -- node[below] {} (stop1);

\end{tikzpicture}
    \end{center}
    \caption{General automated RL policy SBST framework. First, GPT-4 extracts the environment source code from the given environment via Gimitest's functionality. Second, GPT-4 creates the test code which is executed automatically with the trained RL policy. Finally, the test results get outputted.}
    \label{fig:automated_testing}
\end{figure}
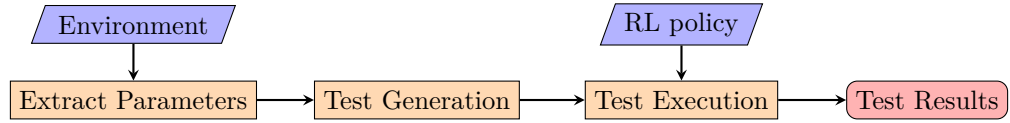

The automated testing pipeline (as described in Figure~\ref{fig:automated_testing}) can also be found in our repository.
First, the GPT-4 analyzes the environment source code. Then, it extracts the essential environment parameters to generate tests that test different environment parameters such as gravity, mass, cart pole length, etc. and then creates the testing code that can be executed to gain test results.
Our experiments indicate that automating RL testing via GPT-4 for the Cartpole environment is possible.

Furthermore, we can automatically analyze via the GPT-4 vision model the failures by, for instance, recording the last frame of the episode, as shown in Figure~\ref{fig:failure_explaination}.
That could be especially interesting when we have a lot of testing data to categorize the different failure types.

\section{Discussion}
Gimitest supports multiple testing methods, including SBST, AT, and MT.
This diversity allows for comprehensive evaluation of RL policies from different angles.
The tool accommodates both turn-based and parallel MARL environments, demonstrating flexibility in handling agent interactions.
Leveraging GPT-4 for automated test code generation and failure analysis streamlines the testing process. It reduces manual effort in extracting environment parameters and generating test cases.
Gimitest facilitates logging and analyzing policy actions, aiding in understanding behavior under various testing conditions.
We conducted the experiments on the ex3 cluster~\citep{cai2021ex3}, which imposes no limitations on running Gimitest.
Besides, the lack of off-the-shelf tools for enabling HPC for RL policy testing prevented us from comparing it with other tools.

\section{Conclusion}
We introduced \emph{Gimitest}, a tool for testing RL policies in environments such as Farama Gymnasium and PettingZoo, which simplifies testing by modifying the reset() and step() methods and allows for customizable tests.
Gimitest can be used for many types of policy testing. It is also available as open source under a permissive licence, making it easily accessible and adaptable for individual needs.

Future extensions of \emph{Gimitest} include supporting parallel execution for faster testing, integrating state-of-the-art explainable RL methods for easier debugging, and exploring automated test code generation using large language models.

\section{Acknowledgment}
Following the journal's policy, we disclose that generative AI technology, including GPT-4, was used to assist in revising this text.

\section*{Conflict of Interest Statement}

The authors declare that the research was conducted in the absence of any commercial or financial relationships that could be construed as a potential conflict of interest.

\section*{Author Contributions}

The paper was written by DG, QM, HS, and AG.
Gimitest was developed by DG and QM. The use case evaluation was implemented by QM and DG.

\section*{Funding}
This work is funded by the EU under grant agreement number 101091783 (MARS Project) and as part of the Horizon Europe HORIZON-CL4-2022-TWIN-TRANSITION-01-03. It is further funded by the Research Council of Norway (project AutoCSP, grant number 324674).

\section*{Data Availability Statement}\label{sec:avail}
\textit{Gimitest} can be downloaded from \url{https://github.com/DennisGross/Gimitest}.

\bibliographystyle{tmlr}
\bibliography{test}

\end{document}